\documentclass[11pt,a4paper]{article}
\usepackage[hyperref]{acl2017}
\usepackage{times}
\usepackage{latexsym}
\usepackage{amsfonts}
\usepackage{algorithm}
\usepackage{algorithmic}

\usepackage{graphicx}
\usepackage{caption}
\usepackage{subcaption}
\usepackage{placeins}

\usepackage{url}

\aclfinalcopy 

\title{Semi-Supervised QA with Generative Domain-Adaptive Nets}

\author{Zhilin Yang ~~ Junjie Hu ~~ Ruslan Salakhutdinov ~~ William W. Cohen \\
  School of Computer Science \\
  Carnegie Mellon University \\
  {\tt \{zhiliny,junjieh,rsalakhu,wcohen\}@cs.cmu.edu}
}

\date{}

\begin{document}
\maketitle
\begin{abstract}
We study the problem of semi-supervised question answering----utilizing unlabeled text to boost the performance of question answering models. We propose a novel training framework, the \textit{Generative Domain-Adaptive Nets}. In this framework, we train a generative model to generate questions based on the unlabeled text, and combine model-generated questions with human-generated questions for training question answering models. We develop novel domain adaptation algorithms, based on reinforcement learning, to alleviate the discrepancy between the model-generated data distribution and the human-generated data distribution. Experiments show that our proposed framework obtains substantial improvement from unlabeled text.
\end{abstract}

\section{Introduction}\label{sec:intro}

Recently, various neural network models were proposed and successfully applied to the tasks of questions answering (QA) and/or reading comprehension \cite{xiong2016dynamic,dhingra2016gated,yang2016words}. While achieving state-of-the-art performance, these models rely on a large amount of labeled data.
However, it is extremely difficult to collect large-scale question answering datasets. Historically, many of the question answering datasets have only thousands of question answering pairs, such as WebQuestions \cite{berant2013semantic}, MCTest \cite{richardson2013mctest}, WikiQA \cite{yang2015wikiqa}, and TREC-QA \cite{voorhees2000building}. Although larger question answering datasets with hundreds of thousands of question-answer pairs have been collected, including SQuAD \cite{rajpurkar2016squad}, MSMARCO \cite{nguyen2016ms}, and NewsQA \cite{trischler2016newsqa}, the data collection process is expensive and time-consuming in practice. This hinders real-world applications for domain-specific question answering.

Compared to obtaining labeled question answer pairs, it is trivial to obtain unlabeled text data. In this work, we study the following problem of semi-supervised question answering: is it possible to leverage unlabeled text to boost the performance of question answering models, especially when only a small amount of labeled data is available? The problem is challenging because conventional manifold-based semi-supervised learning algorithms \cite{zhu2002learning,yang2016revisiting} cannot be straightforwardly applied. Moreover, since the main foci of most question answering tasks are extraction rather than generation, it is also not sensible to use unlabeled text to improve language modeling as in machine translation \cite{gulcehre2015using}.

To better leverage the unlabeled text, we propose a novel neural framework called \textit{Generative Domain-Adaptive Nets} (GDANs). The starting point of our framework is to use linguistic tags to extract possible answer chunks in the unlabeled text, and then train a generative model to generate questions given the answer chunks and their contexts. The model-generated question-answer pairs and the human-generated question-answer pairs can then be combined to train a question answering model, referred to as a \textit{discriminative model} in the following text. However, there is discrepancy between the model-generated data distribution and the human-generated data distribution, which leads to suboptimal discriminative models. To address this issue, we further propose two domain adaptation techniques that treat the model-generated data distribution as a different domain. First, we use an additional \textit{domain tag} to indicate whether a question-answer pair is model-generated or human-generated. We condition the discriminative model on the domain tags so that the discriminative model can learn to factor out domain-specific and domain-invariant representations. Second, we employ a reinforcement learning algorithm to fine-tune the generative model to minimize the loss of the discriminative model in an adversarial way.

In addition, we present a simple and effective baseline method for semi-supervised question answering. Although the baseline method performs worse than our GDAN approach, it is extremely easy to implement and can still lead to substantial improvement when only limited labeled data is available.

We experiment on the SQuAD dataset \cite{rajpurkar2016squad} with various labeling rates and various amounts of unlabeled data. Experimental results show that our GDAN framework consistently improves over both the supervised learning setting and the baseline methods, including adversarial domain adaptation \cite{ganin2014unsupervised} and dual learning \cite{xia2016dual}. More specifically, the GDAN model improves the F1 score by 9.87 points in F1 over the supervised learning setting when 8K labeled question-answer pairs are used.

Our contribution is four-fold. First, different from most of the previous neural network studies on question answering, we study a critical but challenging problem, semi-supervised question answering. Second, we propose the Generative Domain-Adaptive Nets that employ domain adaptation techniques on generative models with reinforcement learning algorithms. Third, we introduce a simple and effective baseline method. Fourth, we empirically show that our framework leads to substantial improvements.

\section{Semi-Supervised Question Answering} \label{sec:qa}

Let us first introduce the problem of \textit{semi-supervised question answering}.

Let $L = \{q^{(i)}, a^{(i)}, p^{(i)}\}_{i = 1}^N$ denote a question answering dataset of $N$ instances, where $q^{(i)}$, $a^{(i)}$, and $p^{(i)}$ are the question, answer, and paragraph of the $i$-th instance respectively.
The goal of question answering is to produce the answer $a^{(i)}$ given the question $q^{(i)}$ along with the paragraph $p^{(i)}$.
We will drop the superscript $\cdot^{(i)}$ when the context is unambiguous. In our formulation, following the setting in SQuAD \cite{rajpurkar2016squad}, we specifically focus on extractive question answering, where $a$ is always a consecutive chunk of text in $p$. More formally, let $p = (p_1, p_2, \cdots, p_T)$ be a sequence of word tokens with $T$ being the length, then $a$ can always be represented as $a = (p_j, p_{j + 1}, \cdots, p_{k - 1}, p_k)$, where $j$ and $k$ are the start and end token indices respectively. The questions can also be represented as a sequence of word tokens $q = (q_1, q_2, \cdots, q_{T'})$ with length $T'$.

In addition to the labeled dataset $L$, in the semi-supervised setting, we are also given a set of unlabeled data, denoted as $U = \{a^{(i)}, p^{(i)}\}_{i = 1}^M$, where $M$ is the number of unlabeled instances. Note that it is usually trivial to have access to an almost infinite number of paragraphs $p$ from sources such as Wikipedia articles and other web pages. And since the answer $a$ is always a consecutive chunk in $p$, we argue that it is also sensible to extract possible answer chunks from the unlabeled text using linguistic tags. We will discuss the technical details of answer chunk extraction in Section \ref{sec:extract}, and in the formulation of our framework, we assume that the answer chunks $a$ are available.

Given both the labeled data $L$ and the unlabeled data $U$, the goal of semi-supervised question answering is to learn a question answering model $D$ that captures the probability distribution $\mathbb{P}(a | p, q)$. We refer to this question answering model $D$ as the \textit{discriminative model}, in contrast to the generative model that we will present in Section \ref{sec:gen}.

\subsection{A Simple Baseline} \label{sec:baseline}

We now present a simple baseline for semi-supervised question answering. Given a paragraph $p = (p_1, p_2, \cdots, p_T)$ and the answer $a = (p_j, p_{j + 1}, \cdots, p_{k - 1}, p_k)$, we extract $(p_{j - W}, p_{j - W + 1}, \cdots, p_{j - 1}, p_{k + 1}, p_{k + 2}, p_{k + W})$ from the paragraph and treat it as the question. Here $W$ is the window size and is set at 5 in our experiments so that the lengths of the questions are similar to human-generated questions. The context-based question-answer pairs on $U$ are combined with human-generated pairs on $L$ for training the discriminative model. Intuitively, this method extracts the contexts around the answer chunks to serve as hints for the question answering model. Surprisingly, this simple baseline method leads to substantial improvements when labeled data is limited.

\section{Generative Domain-Adaptive Nets}\label{sec:model}

Though the simple method described in Section \ref{sec:baseline} can lead to substantial improvement, we aim to design a learning-based model to move even further. In this section, we will describe the model architecture and the training algorithms for the GDANs. We will use a notation in the context of question answering following Section \ref{sec:qa}, but one should be able to extend the notion of GDANs to other applications as well.

The GDAN framework consists of two models, \textit{a discriminative model} and a \textit{generative model}. We will first discuss the two models in detail in the context of question answering, and then present an algorithm based on reinforcement learning to combine the two models.

\subsection{Discriminative Model}

The discriminative model learns the conditional probability of an answer chunk given the paragraph and the question, i.e., $\mathbb{P}(a | p, q)$. We employ a gated-attention (GA) reader \cite{dhingra2016gated} as our base model in this work, but our framework does not make any assumptions about the base models being used. The discriminative model is referred to as $D$.

The GA model consists of $K$ layers with $K$ being a hyper-parameter. Let $\mathbf{H}_p^k$ be the intermediate paragraph representation at layer $k$, and $\mathbf{H}_q$ be the question representation. The paragraph representation $\mathbf{H}_p^k$ is a $T \times d$ matrix, and the question representation $\mathbf{H}_q$ is a $T' \times d$ matrix, where $d$ is the dimensionality of the representations. Given the paragraph $p$, we apply a bidirectional Gated Recurrent Unit (GRU) network \cite{chung2014empirical} on top of the embeddings of the sequence $(p_1, p_2, \cdots, p_T)$, and obtain the initial paragraph representation $\mathbf{H}_p^0$. Given the question $q$, we also apply another bidirectional GRU to obtain the question representation $\mathbf{H}_q$.

The question and paragraph representations are combined with the gated-attention (GA) mechanism \cite{dhingra2016gated}. More specifically, for each paragraph token $p_i$, we compute
\[
\alpha_j = \frac{\exp \mathbf{h}_{q, j}^T \mathbf{h}_{p, i}^{k - 1}}{\sum_{j' = 1}^{T'} \exp \mathbf{h}_{q, j'}^T \mathbf{h}_{p, i}^{k - 1}}
\]
\[
\mathbf{h}_{p, i}^k = \sum_{j = 1}^{T'} \alpha_j \mathbf{h}_{q, j} \odot \mathbf{h}_{p, i}^{k - 1}
\]
where $\mathbf{h}_{p, i}^k$ is the $i$-th row of $\mathbf{H}_p^k$ and $\mathbf{h}_{q, j}$ is the $j$-th row of $\mathbf{H}_q$.

Since the answer $a$ is a sequence of consecutive word tokens in the paragraph $p$, we apply two softmax layers on top of $\mathbf{H}_p^K$ to predict the start and end indices of $a$, following Yang et al. \shortcite{yang2016words}.

\subsubsection{Domain Adaptation with Tags} \label{sec:tag}

We will train our discriminative model on both model-generated question-answer pairs and human-generated pairs. However, even a well-trained generative model will produce questions somewhat different from human-generated ones. Learning from both human-generated data and model-generated data can thus lead to a biased model.
To alleviate this issue, we propose to view the model-generated data distribution and the human-generated data distribution as two different data domains and explicitly incorporate domain adaptation into the discriminative model.

More specifically, we use a \textit{domain tag} as an additional input to the discriminative model. We use the tag ``d\_true'' to represent the domain of human-generated data (i.e., the \textit{true} data), and ``d\_gen'' for the domain of model-generated data. Following a practice in domain adaptation \cite{johnson2016google,chu2017empirical}, we append the domain tag to the end of both the questions and the paragraphs. By introducing the domain tags, we expect the discriminative model to factor out domain-specific and domain-invariant representations. At test time, the tag ``d\_true'' is appended.

\subsection{Generative Model} \label{sec:gen}

The generative model learns the conditional probability of generating a question given the paragraph and the answer, i.e., $\mathbb{P}(q | p, a)$. We implement the generative model as a sequence-to-sequence model \cite{sutskever2014sequence} with a copy mechanism \cite{gu2016incorporating,gulcehre2016pointing}.

The generative model consists of an encoder and a decoder. An encoder is a GRU that encodes the input paragraph into a sequence of hidden states $\mathbf{H}$. We inject the answer information by appending an additional zero/one feature to the word embeddings of the paragraph tokens; i.e., if a word token appears in the answer, the feature is set at one, otherwise zero.

The decoder is another GRU with an attention mechanism over the encoder hidden states $\mathbf{H}$. At each time step, the generation probabilities over all word types are defined with a copy mechanism:
\begin{equation} \label{eq:p}
\mathbf{p}_{\mbox{overall}} = g_t \mathbf{p}_{\mbox{vocab}} + (1 - g_t) \mathbf{p}_{\mbox{copy}}
\end{equation}
where $g_t$ is the probability of generating the token from the vocabulary, while $(1 - g_t)$ is the probability of copying a token from the paragraph. The probability $g_t$ is computed based on the current hidden state $\mathbf{h}_t$:
\[
g_t = \sigma (\mathbf{w}_g^T \mathbf{h}_t)
\]
where $\sigma$ denotes the logistic function and $\mathbf{w}_g$ is a vector of model parameters. The generation probabilities $\mathbf{p}_{\mbox{vocab}}$ are defined as a softmax function over the word types in the vocabulary, and the copying probabilities $\mathbf{p}_{\mbox{copy}}$ are defined as a softmax function over the word types in the paragraph. Both $\mathbf{p}_{\mbox{vocab}}$ and $\mathbf{p}_{\mbox{copy}}$ are defined as a function of the current hidden state $\mathbf{h}_t$ and the attention results \cite{gu2016incorporating}.

\subsection{Training Algorithm}

\begin{algorithm}[t]
    \caption{Training Generative Domain-Adaptive Nets}
    \label{algo}
\begin{algorithmic}
    \STATE {\bfseries Input:} labeled data $L$, unlabeled data $U$, \#iterations $T_G$ and $T_D$
    \STATE Initialize $G$ by MLE training on $L$
    \STATE Randomly initialize $D$
    \WHILE {not stopping}
        \FOR {$t \gets 1 \mbox{~to~} T_D$} {
            \STATE Update $D$ to maximize $J(L, \mbox{d\_true}, D) + J(U_G, \mbox{d\_gen}, D)$ with SGD
        }
        \ENDFOR
        \FOR {$t \gets 1 \mbox{~to~} T_G$} {
            \STATE Update $G$ to maximize $J(U_G, \mbox{d\_true}, D)$ with Reinforce and SGD
        }
        \ENDFOR
    \ENDWHILE
    \STATE {\bfseries return} model $D$
\end{algorithmic}
\end{algorithm}

\begin{figure*}[t]
    \centering
    \begin{subfigure}[b]{0.3\textwidth}
        \includegraphics[width=\textwidth]{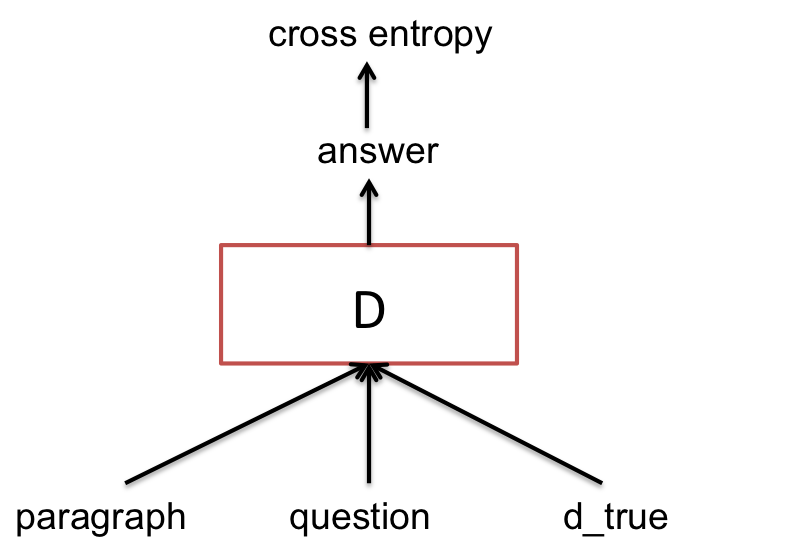}
        \caption{Training the discriminative model on labeled data.}
    \end{subfigure}
    ~ 
    \begin{subfigure}[b]{0.3\textwidth}
        \includegraphics[width=\textwidth]{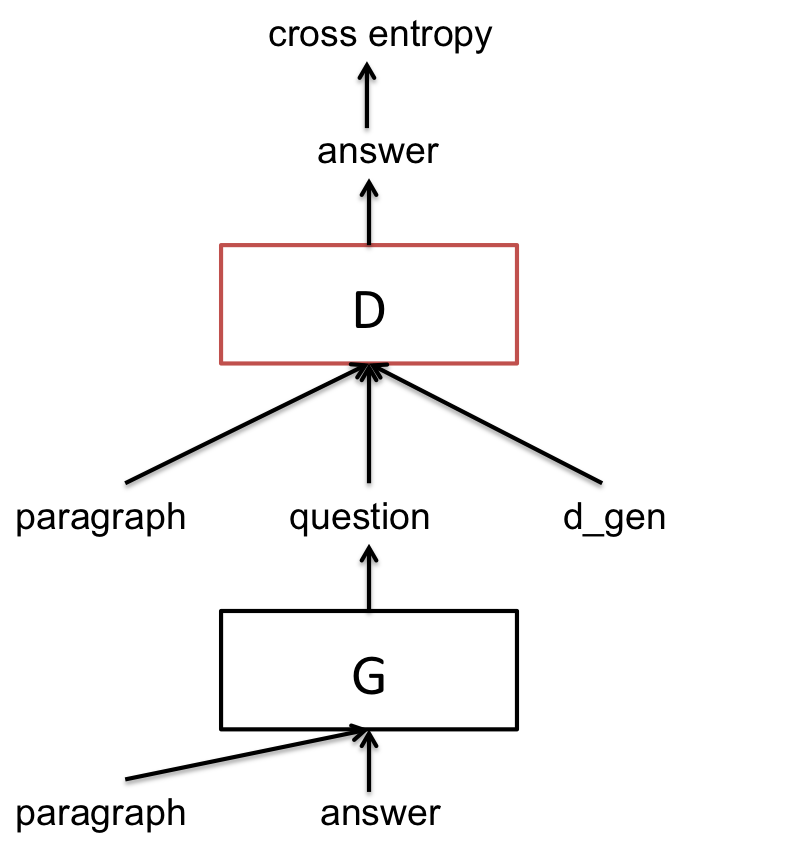}
        \caption{Training the discriminative model on unlabeled data.}
    \end{subfigure}
    ~ 
    \begin{subfigure}[b]{0.3\textwidth}
        \includegraphics[width=\textwidth]{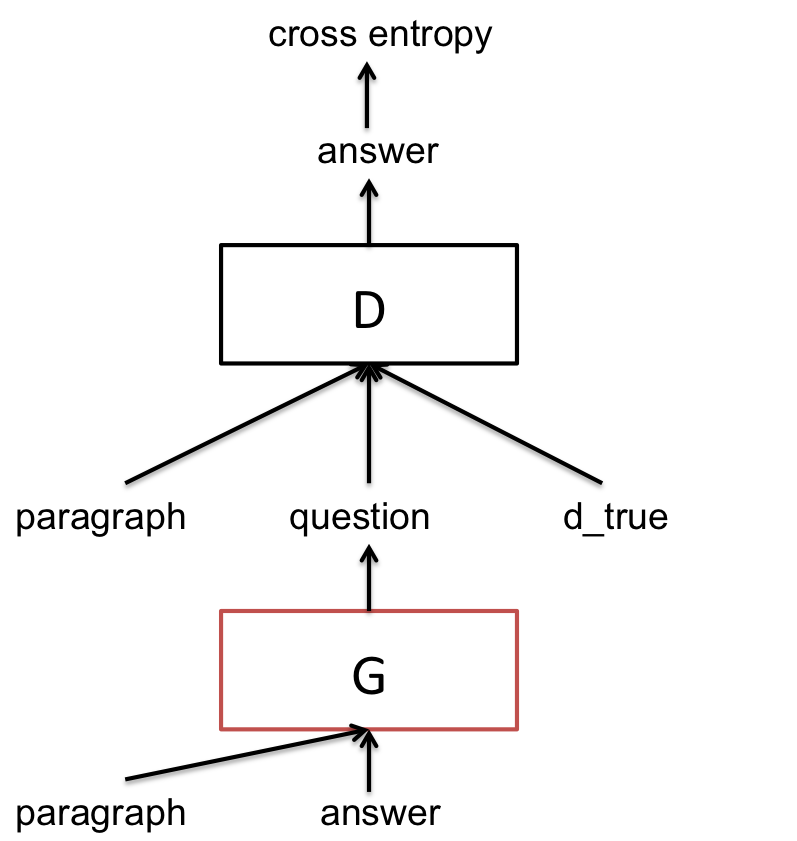}
        \caption{Training the generative model on unlabeled data.}
    \end{subfigure}
    \caption{\small Model architecture and training. Red boxes denote the modules being updated. ``d\_true'' and ``d\_gen'' are two domain tags. $D$ is the discriminative model and $G$ is the generative model. The objectives for the three cases are all to minimize the cross entropy loss of the answer chunks.}\label{fig:arch}
\end{figure*}

We first define the objective function of the GDANs, and then present an algorithm to optimize the given objective function. Similar to the Generative Adversarial Nets (GANs) \cite{goodfellow2014generative} and adversarial domain adaptation \cite{ganin2014unsupervised}, the discriminative model and the generative model have different objectives in our framework. However, rather than formulating the objective as an adversarial game between the two models \cite{goodfellow2014generative,ganin2014unsupervised}, in our framework, the discriminative model relies on the data generated by the generative model, while the generative model aims to match the model-generated data distribution with the human-generated data distribution using the signals from the discriminative model.

Given a labeled dataset $L = \{p^{(i)}, q^{(i)}, a^{(i)}\}_{i = 1}^N$, the objective function of a discriminative model $D$ for a supervised learning setting can be written as $\sum_{p^{(i)}, q^{(i)}, a^{(i)} \in L} \log \mathbb{P}_D(a^{(i)} | p^{(i)}, q^{(i)})$, where $\mathbb{P}_D$ is a probability distribution defined by the model $D$. Since we also incorporate domain tags into the model $D$, we denote the objective function as

\small
\[
J(L, \mbox{tag}, D) = \frac{1}{|L|} \sum_{p^{(i)}, q^{(i)}, a^{(i)} \in L} \log \mathbb{P}_{D, \mbox{tag}}(a^{(i)} | p^{(i)}, q^{(i)})
\]
\normalsize
meaning that the domain tag, ``tag'', is appended to the dataset $L$. We use $|L| = N$ to denote the number of the instances in the dataset $L$. The objective function is averaged over all instances such that we can balance labeled and unlabeled data.

Let $U_G$ denote the dataset obtained by generating questions on the unlabeled dataset $U$ with the generative model $G$. The objective of the discriminative model is then to maximize $J$ for both labeled and unlabeled data under the domain adaptation notions, i.e., $J(L, \mbox{d\_true}, D) + J(U_G, \mbox{d\_gen}, D)$.

Now we discuss the objective of the generative model. Similar to the dual learning \cite{xia2016dual} framework, one can define an auto-encoder objective. In this case, the generative model aims to generate questions that can be reconstructed by the discriminative model, i.e., maximizing $J(U_G, \mbox{d\_gen}, D)$. However, this objective function can lead to degenerate solutions because the questions can be thought of as an overcomplete representation of the answers \cite{vincent2010stacked}. For example, given $p$ and $a$, the generative model might learn to generate trivial questions such as copying the answers, which does not contributed to learning a better $D$.

Instead, we leverage the discriminative model to better match the model-generated data distribution with the human-generated data distribution. We propose to define an adversarial training objective $J(U_G, \mbox{d\_true}, D)$. We append the tag ``d\_true'' instead of ``d\_gen'' for the model-generated data to ``fool'' the discriminative model. 
Intuitively, the goal of G is to generate "useful" questions where the usefulness is measured by the probability that the generated questions can be answered correctly by $D$.


The overall objective function now can be written as
\begin{eqnarray*}
&\max_D& J(L, \mbox{d\_true}, D) + J(U_G, \mbox{d\_gen}, D) \\
&\max_G& J(U_G, \mbox{d\_true}, D)
\end{eqnarray*}

With the above objective function in mind, we present a training algorithm in Algorithm \ref{algo} to train a GDAN. We first pretrain the generative model on the labeled data $L$ with maximum likelihood estimation (MLE):
\[
\max_G \sum_{i = 1}^N \sum_{t = 1}^{T'} \log \mathbb{P}_G(q^{(i)}_t | q^{(i)}_{< t}, p^{(i)}, a^{(i)})
\]
where $\mathbb{P}_G$ is the probability defined by Eq. \ref{eq:p}.

We then alternatively update $D$ and $G$ based on their objectives. To update $D$, we sample one batch from the labeled data $L$ and one batch from the unlabeled data $U_G$, and combine the two batches to perform a gradient update step. Since the output of $G$ is discrete and non-differentiable, we use the Reinforce algorithm \cite{williams1992simple} to update $G$. The action space is all possible questions with length $T'$ (possibly with padding) and the reward is the objective function $J(U_G, \mbox{d\_true}, D)$. Let $\theta_G$ be the parameters of $G$. The gradient can be written as

\small
\begin{eqnarray*}
&& \frac{\partial J(U_G, \mbox{d\_true}, D)}{\partial \theta_G} \\
&=& \mathbb{E}_{\mathbb{P}_G(q | p, a)} (\log \mathbb{P}_{D, \mbox{d\_true}}(a | p, q) - b) \frac{\partial \log \mathbb{P}_G(q | p, a)}{\partial \theta_G}
\end{eqnarray*}
\normalsize

\noindent where we use an average reward from samples as the baseline $b$. We approximate the expectation $\mathbb{E}_{\mathbb{P}_G(q | p, a)}$ by sampling one instance at a time from $\mathbb{P}_G(q | p, a)$ and then do an update step. This training algorithm is referred to as reinforcement learning (RL) training in the following sections.
The overall architecture and training algorithm are illustrated in Figure \ref{fig:arch}.

\textbf{MLE vs RL.} The generator $G$ has two training phases--MLE training and RL training, which are different in that: 1) RL training does not require labels, so $G$ can explore a broader data domain of $p$ using unlabeled data, while MLE training requires labels; 2) MLE maximizes $\log P(q | p, a)$, while RL maximizes $\log P_D(a | q, p)$. Since $\log P(q | a, p)$ is the sum of $\log P(q | p)$ and $\log P(a | q, p)$ (plus a constant), maximizing $\log P(a | q, p)$ does not require modeling $\log P(q | p)$ that is irrelevant to QA, which makes optimization easier. Moreover, maximizing $\log P(a | q, p)$ is consistent with the goal of QA.






\begin{table*}[t]
\caption{\label{tab:sample} \small Sampled generated questions given the paragraphs and the answers. \textit{P} means paragraphs, \textit{A} means answers, \textit{GQ} means groundtruth questions, and \textit{Q} means questions generated by our models. \textit{MLE} refers to maximum likelihood training, and \textit{RL} refers to reinforcement learning so as to maximize $J(U_G, \mbox{d\_true}, D)$. We truncate the paragraphs to only show tokens around the answer spans with a window size of 20.}
\small
\begin{center}
\begin{tabular}{p{15cm}}
\hline
\textbf{P1:} is mediated by ige , which triggers degranulation of mast cells and basophils when cross - linked by antigen . type ii hypersensitivity occurs when antibodies bind to antigens on the patient ' s own cells , marking them for destruction . this \\
\textbf{A:} type ii hypersensitivity \\
\textbf{GQ:} antibody - dependent hypersensitivity belongs to what class of hypersensitivity ? \\
\textbf{Q (MLE):} what was the UNK of the patient ' s own cells ? \\
\textbf{Q (RL):} what occurs when antibodies bind to antigens on the patient ' s own cells by antigen when cross \\
\hline
\textbf{P2:} an additional warming of the earth ' s surface . they calculate with confidence that co0 has been responsible for over half the enhanced greenhouse effect . they predict that under a `` business as usual " ( bau ) scenario , \\
\textbf{A:} over half \\
\textbf{GQ:} how much of the greenhouse effect is due to carbon dioxide ? \\
\textbf{Q (MLE):} what is the enhanced greenhouse effect ? \\
\textbf{Q (RL):} what the enhanced greenhouse effect that co0 been responsible for \\
\hline
\textbf{P3:} ) narrow gauge lines , which are the remnants of five formerly government - owned lines which were built in mountainous areas . \\
\textbf{A:} mountainous areas \\
\textbf{GQ:} where were the narrow gauge rail lines built in victoria ? \\
\textbf{Q (MLE):} what is the government government government - owned lines built ? \\
\textbf{Q (RL):} what were the remnants of government - owned lines built in \\
\hline
\textbf{P4:} but not both ). in 0000 , bankamericard was renamed and spun off into a separate company known today as visa inc . \\
\textbf{A:} visa inc . \\
\textbf{GQ:} what present - day company did bankamericard turn into ? \\
\textbf{Q (MLE):} what was the separate company bankamericard ? \\
\textbf{Q (RL):} what today as bankamericard off into a separate company known today as spun off into a separate company known today \\
\hline
\textbf{P5:} legrande writes that " the formulation of a single all - encompassing definition of the term is extremely difficult , if \\
\textbf{A:} legrande \\
\textbf{GQ:} who wrote that it is difficult to produce an all inclusive definition of civil disobedience ? \\
\textbf{Q (MLE):} what is the term of a single all all all all encompassing definition of a single all \\
\textbf{Q (RL):} what writes " the formulation of a single all - encompassing definition of the term all encompassing encompassing encompassing encompassing \\
\end{tabular}
\end{center}
\normalsize
\end{table*}

\section{Experiments}\label{sec:exp}

\subsection{Answer Extraction} \label{sec:extract}

As discussed in Section \ref{sec:qa}, our model assumes that answers are available for unlabeled data. In this section, we introduce how we use linguistic tags and rules to extract answer chunks from unlabeled text.

To extract answers from massive unlabelled Wikipedia articles, we first sample 205,511 Wikipedia articles that are not used in the training, development and test sets in the SQuAD dataset.  We extract the paragraphs from each article, and limit the length of each paragraph at the word level to be less than 850. In total, we obtain 950,612 paragraphs from unlabelled articles.

Answers in the SQuAD dataset can be categorized into ten types, i.e., ``Date'', ``Other Numeric'', ``Person'', ``Location'', ``Other Entity'', ``Common Noun Phrase'', ``Adjective Phrase'', ``Verb Phrase'', ``Clause'' and ``Other'' \cite{rajpurkar2016squad}. For each paragraph from the unlabeled articles, we utilize Stanford Part-Of-Speech (POS) tagger~\cite{toutanova2003feature} to label each word with the corresponding POS tag, and implement a simple constituency parser to extract the noun phrase, verb phrase, adjective and clause based on a small set of constituency grammars. Next, we use Stanford Named Entity Recognizer (NER)~\cite{finkel2005incorporating} to assign each word with one of the seven labels, i.e., ``Date'', ``Money'', ``Percent'', ``location'', ``Organization'' and ``Time''.  We then categorize a span of consecutive words with the same NER tags of either ``Money'' or ``Percent'' as the answer of the type ``Other Numeric''.  Similarly, we categorize a span of consecutive words with the same NER tags of ``Organization'' as the answer of the type ``Other Entity''.  Finally, we subsample five answers from all the extracted answers for each paragraph according to the percentage of answer types in the SQuAD dataset. We obtain 4,753,060 answers in total, which is about 50 times larger than the number of answers in the SQuAD dataset.

\subsection{Settings and Comparison Methods}

The original SQuAD dataset consists of 87,636 training instances and 10,600 development instances. Since the test set is not published, we split 10\% of the training set as the test set, and the remaining 90\% serves as the actual training set. Instances are split based on articles; i.e., paragraphs in one article always appear in only one set. We tune the hyper-parameters and perform early stopping on the development set using the F1 scores, and the performance is evaluated on the test set using both F1 scores and exact matching (EM) scores \cite{rajpurkar2016squad}.

We compare the following methods. \textbf{SL} is the supervised learning setting where we train the model $D$ solely on the labeled data $L$. \textbf{Context} is the simple context-based method described in Section \ref{sec:baseline}. \textbf{Context + domain} is the ``Context'' method with domain tags as described in Section \ref{sec:tag}. \textbf{Gen} is to train a generative model and use the generated questions as additional training data. \textbf{Gen + GAN} refers to the domain adaptation method using GANs \cite{ganin2014unsupervised}; in contrast to the original work, the generative model is updated using Reinforce. \textbf{Gen + dual} refers to the dual learning method \cite{xia2016dual}. \textbf{Gen + domain} is ``Gen'' with domain tags, while the generative model is trained with MLE and fixed. \textbf{Gen + domain + adv} is the approach we propose (Cf. Figure \ref{fig:arch} and Algorithm \ref{algo}), with ``adv'' meaning adversarial training based on Reinforce. We use our own implementation of ``Gen + GAN'' and ``Gen + dual'', since the GAN model \cite{ganin2014unsupervised} does not handle discrete features and the dual learning model \cite{xia2016dual} cannot be directly applied to question answering. When implementing these two baselines, we adopt the learning schedule introduced by Ganin and Lempitsky \shortcite{ganin2014unsupervised}, i.e., gradually increasing the weights of the gradients for the generative model $G$.

\subsection{Results and Analysis}

We study the performance of different models with varying labeling rates and unlabeled dataset sizes. Labeling rates are the percentage of training instances that are used to train $D$. The results are reported in Table \ref{tab:res}. Though the unlabeled dataset we collect consists of around 5 million instances, we also sample a subset of around 50,000 instances to evaluate the effects of the size of unlabeled data. The highest labeling rate in Table \ref{tab:res} is $0.9$ because 10\% of the training instances are used for testing. Since we do early stopping on the development set using the F1 scores, we also report the development F1. We report two metrics, the F1 scores and the exact matching (EM) scores \cite{rajpurkar2016squad}, on the test set. All metrics are computed using the official evaluation scripts.

\textbf{SL v.s. SSL.} We observe that semi-supervised learning leads to consistent improvements over supervised learning in all cases. Such improvements are substantial when labeled data is limited. For example, the GDANs improve over supervised learning by 9.87 points in F1 and 7.26 points in EM when the labeling rate is $0.1$. With our semi-supervised learning approach, we can use only $0.1$ training instances to obtain even better performance than a supervised learning approach with $0.2$ training instances, saving more than half of the labeling costs.

\textbf{Comparison with Baselines.} By comparing ``Gen + domain + adv'' with ``Gen + GAN'' and ``Gen + Dual'', it is clear that the GDANs perform substantially better than GANs and dual learning. With labeling rate $0.1$, GDANs outperform dual learning and GANs by 2.47 and 4.29 points respectively in terms of F1.

\textbf{Ablation Study.} We also perform an ablation study by examining the effects of ``domain'' and ``adv'' when added to ``gen''. It can be seen that both the domain tags and the adversarial training contribute to the performance of the GDANs when the labeling rate is equal to or less than $0.5$. With labeling rate $0.9$, adding domain tags still leads to better performance but adversarial training does not seem to improve the performance by much.

\textbf{Unlabeled Data Size.} Moreover, we observe that the performance can be further improved when a larger unlabeled dataset is used, though the gain is relatively less significant compared to changing the model architectures. For example, increasing the unlabeled dataset size from 50K to 5M, the performance of GDANs increases by 0.38 points in F1 and 0.52 points in EM.

\textbf{Context-Based Method.} Surprisingly, the simple context-based method, though performing worse than GDANs, still leads to substantial gains; e.g., 7.00 points in F1 with labeling rate $0.1$. Adding domain tags can improve the performance of the context-based method as well.

\begin{figure}[t]
    \centering
    \includegraphics[width=0.4\textwidth]{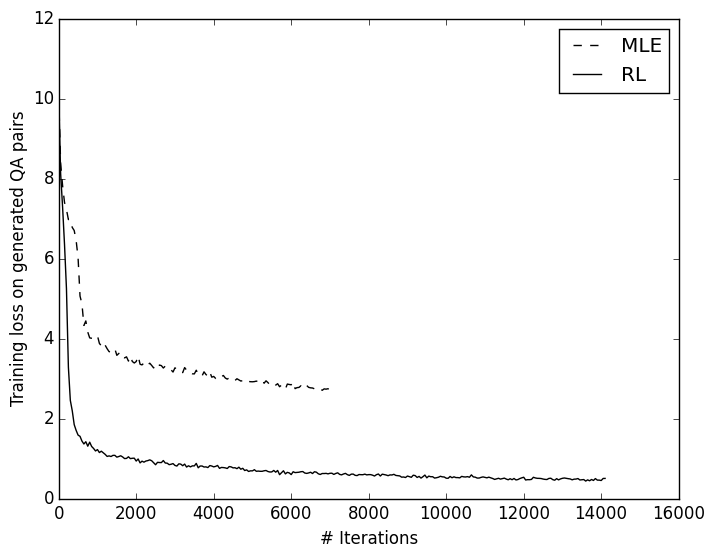}
    \caption{\small Comparison of discriminator training loss $- J(U_G, \mbox{d\_gen}, D)$ on generated QA pairs. The lower the better. MLE refers to questions generated by maximum likelihood training, and RL refers to questions generated by reinforcement learning.}\label{fig:gen}
\end{figure}

\textbf{MLE vs RL.} We plot the loss curve of $- J(U_G, \mbox{d\_gen}, D)$ for both the MLE-trained generator (``Gen + domain'') and the RL-trained generator (``Gen + domain + adv'') in Figure \ref{fig:gen}. We observe that the training loss for D on RL-generated questions is lower than MLE-generated questions, which confirms that RL training maximizes $\log P(a | p, q)$.

\textbf{Samples of Generated Questions.} We present some questions generated by our model in Table \ref{tab:sample}. The generated questions are post-processed by removing repeated subs-sequences. Compared to MLE-generated questions, RL-generated questions are more informative (Cf., P1, P2, and P4), and contain less ``UNK'' (unknown) tokens (Cf., P1). Moreover, both semantically and syntactically, RL-generated questions are more accurate (Cf., P3 and P5).

\begin{table*}[h]
\caption{\label{tab:res} \small Performance with various labeling rates, unlabeled data sizes $|U|$, and methods. ``Dev'' denotes the development set, and ``test'' denotes the test set. F1 and EM are two metrics.}
\begin{center}
\begin{tabular}{llllll}
Labeling rate & $|U|$ & Method & Dev F1 & Test F1 & Test EM \\ \hline \\
0.1 & 50K & SL & 0.4262 & 0.3815 & 0.2492 \\
0.1 & 50K & Context & 0.5046 & 0.4515 & 0.2966 \\
0.1 & 50K & Context + domain & 0.5139 & 0.4575 & 0.3036 \\
0.1 & 50K & Gen & 0.5049 & 0.4553 & 0.3018 \\
0.1 & 50K & Gen + GAN & 0.4897 & 0.4373 & 0.2885 \\
0.1 & 50K & Gen + dual & 0.5036 & 0.4555 & 0.3005 \\
0.1 & 50K & Gen + domain & 0.5234 & 0.4703 & 0.3145 \\
0.1 & 50K & Gen + domain + adv & \textbf{0.5313} & \textbf{0.4802} & \textbf{0.3218} \\ \hline
0.2 & 50K & SL & 0.5134 & 0.4674 & 0.3163 \\
0.2 & 50K & Context & 0.5652 & 0.5132 & 0.3573 \\
0.2 & 50K & Context + domain & 0.5672 & 0.5200 & 0.3581 \\
0.2 & 50K & Gen & 0.5643 & 0.5159 & 0.3618 \\
0.2 & 50K & Gen + GAN & 0.5525 & 0.5037 & 0.3470 \\
0.2 & 50K & Gen + dual & 0.5720 & 0.5192 & 0.3612 \\
0.2 & 50K & Gen + domain & 0.5749 & 0.5216 & 0.3658 \\
0.2 & 50K & Gen + domain + adv & \textbf{0.5867} & \textbf{0.5394} & \textbf{0.3781} \\ \hline
0.5 & 50K & SL & 0.6280 & 0.5722 & 0.4187 \\
0.5 & 50K & Context & 0.6300 & 0.5740 & 0.4195 \\
0.5 & 50K & Context + domain & 0.6307 & 0.5791 & 0.4237 \\
0.5 & 50K & Gen & 0.6237 & 0.5717 & 0.4155 \\
0.5 & 50K & Gen + GAN & 0.6110 & 0.5590 & 0.4044 \\
0.5 & 50K & Gen + dual & 0.6368 & 0.5746 & 0.4163 \\
0.5 & 50K & Gen + domain & \textbf{0.6378} & 0.5826 & 0.4261 \\
0.5 & 50K & Gen + domain + adv & 0.6375 & \textbf{0.5831} & \textbf{0.4267} \\ \hline
0.9 & 50K & SL & \textbf{0.6611} & 0.6070 & 0.4534 \\
0.9 & 50K & Context & 0.6560 & 0.6028 & 0.4507 \\
0.9 & 50K & Context + domain & 0.6553 & \textbf{0.6105} & 0.4557 \\
0.9 & 50K & Gen & 0.6464 & 0.5970 & 0.4445 \\
0.9 & 50K & Gen + GAN & 0.6396 & 0.5874 & 0.4317 \\
0.9 & 50K & Gen + dual & 0.6511 & 0.5892 & 0.4340 \\
0.9 & 50K & Gen + domain & \textbf{0.6611} & 0.6102 & \textbf{0.4573} \\
0.9 & 50K & Gen + domain + adv & 0.6585 & 0.6043 & 0.4497 \\ \hline
0.1 & 5M & SL & 0.4262 & 0.3815 & 0.2492 \\
0.1 & 5M & Context & 0.5140 & 0.4641 & 0.3014 \\
0.1 & 5M & Context + domain & 0.5166 & 0.4599 & 0.3083 \\
0.1 & 5M & Gen & 0.5099 & 0.4619 & 0.3103 \\
0.1 & 5M & Gen + domain & 0.5301 & 0.4703 & 0.3227 \\
0.1 & 5M & Gen + domain + adv & \textbf{0.5442} & \textbf{0.4840} & \textbf{0.3270} \\ \hline
0.9 & 5M & SL & 0.6611 & 0.6070 & 0.4534 \\
0.9 & 5M & Context & 0.6605 & 0.6026 & 0.4473 \\
0.9 & 5M & Context + domain & 0.6642 & 0.6066 & 0.4548 \\
0.9 & 5M & Gen & 0.6647 & 0.6065 & \textbf{0.4600} \\
0.9 & 5M & Gen + domain & \textbf{0.6726} & 0.6092 & 0.4599 \\
0.9 & 5M & Gen + domain + adv & 0.6670 & \textbf{0.6102} & 0.4531 \\
\end{tabular}
\end{center}
\end{table*}

\section{Related Work}\label{sec:related}

\textbf{Semi-Supervised Learning.}  Semi-supervised learning has been extensively studied in literature~\cite{zhu2005semi}. A batch of novel models have been recently proposed for semi-supervised learning based on representation learning techniques, such as generative models \cite{kingma2014semi}, ladder networks \cite{rasmus2015semi} and graph embeddings \cite{yang2016revisiting}. However, most of the semi-supervised learning methods are based on combinations of the supervised loss $p(\mathbf{y} | \mathbf{x})$ and an unsupervised loss $p(\mathbf{x})$. In the context of reading comprehension, directly modeling the likelihood of a paragraph would not possibly improve the supervised task of question answering. Moreover, traditional graph-based semi-supervised learning~\cite{zhu2002learning} cannot be easily extended to modeling the unlabeled answer chunks.

\textbf{Domain Adaptation.} Domain adaptation has been successfully applied to various tasks, such as classification~\cite{ganin2014unsupervised} and machine translation~\cite{johnson2016google, chu2017empirical}.  Several techniques on domain adaptation~\cite{glorot2011domain} focus on learning distribution invariant features by sharing the intermediate representations for downstream tasks.
Another line of research on domain adaptation attempt to match the distance between different domain distributions in a low dimensional space~\cite{long2015learning, baktashmotlagh2013unsupervised}. There are also methods seeking a domain transition from the source domain to the target domain \cite{gong2012geodesic, gopalan2011domain, pan2011domain}. Our work gets inspiration from a practice in Johnson et al. \shortcite{johnson2016google} and Chu et al. \shortcite{chu2017empirical} based on appending domain tags.
However, our method is different from the above methods in that we apply domain adaptation techniques to the outputs of a generative model rather than a natural data domain.

\textbf{Question Answering.} 
Various neural models based on attention mechanisms~\cite{wang2016machine,seo2016bidirectional,xiong2016dynamic,wang2016multi,dhingra2016gated,kadlec2016text,trischler2016natural,sordoni2016iterative,cui2016attention,chen2016thorough} have been proposed to tackle the tasks of question answering and reading comprehension.  However, the performance of these neural models largely relies on a large amount of labeled data available for training.

\textbf{Learning with Multiple Models.} GANs \cite{goodfellow2014generative} formulated a adversarial game between a discriminative model and a generative model for generating realistic images. Ganin and Lempitsky \cite{ganin2014unsupervised} employed a similar idea to use two models for domain adaptation. Review networks \cite{yang2016review} employ a discriminative model as a regularizer for training a generative model. In the context of machine translation, given a language pair, various recent work studied jointly training models to learn the mappings in both directions \cite{tu2016modeling,xia2016dual}.

\section{Conclusions} \label{sec:conc}

We study a critical and challenging problem, semi-supervised question answering. We propose a novel neural framework called Generative Domain-Adaptive Nets, which incorporate domain adaptation techniques in combination with generative models for semi-supervised learning. Empirically, we show that our approach leads to substantial improvements over supervised learning models and outperforms several strong baselines including GANs and dual learning. In the future, we plan to apply our approach to more question answering datasets in different domains. It will also be intriguing to generalize GDANs to other applications.

\small
\textbf{Acknowledgements.}
This work was funded by the Office of Naval Research grants N000141512791 and N000141310721 and NVIDIA. 
\normalsize

\FloatBarrier

\bibliography{acl2017}
\bibliographystyle{acl_natbib}

\end{document}